\documentclass[preprint,12pt]{elsarticle}

\usepackage{amssymb}
\usepackage{amsmath, color}

\usepackage{caption}
\usepackage{subcaption}
\usepackage{graphicx}
\usepackage{booktabs}
\usepackage{array}
\usepackage{float}

\journal{ISPRS Journal of Photogrammetry and Remote Sensing}

\begin{document}

\begin{frontmatter}

\title{EcoVision: AI-Powered Drone Imaging for Salt Marsh Vegetation Monitoring and Dominance Mapping}

\author[label1]{Innocent Onyenonachi}
\author[label2]{Peter J. Lawerance}
\author[label1]{ Nadia Kanwal}

\affiliation[label1]{organization={School of Computer Science and Mathematics},
            addressline={Keele University},
            postcode={ST5 3BG}, 
            country={United Kingdom}}
\affiliation[label2]{organization={School of Life Science},
            addressline={Keele University},
            postcode={ST5 3BG}, 
            country={United Kingdom}}

\begin{abstract}
High-resolution RGB imagery acquired from low-altitude UAV surveys was processed through a modular pipeline incorporating transformer-based semantic segmentation, connected-component vegetation extraction, fine-grained species classification using a ConvNeXt architecture, and grid-based dominance scoring at 2$\times$2m resolution. The framework targeted two ecologically significant halophytic grasses, Spartina maritima and Puccinellia maritima, and was trained using a curated and manually annotated UAV imagery, along with biodiversity imagery sourced from publicly accessible datasets.
In order to identify these plants from the imagery, our segmentation yielded reliable species masks (mean IoU $\approx$ 0.56; pixel-level accuracy $\approx$ 0.96), while object-level classification achieved very good discrimination (F1 $\approx$ 0.99). Dominance estimates closely matched quadrat-based field surveys, with mean absolute differences below 8\%, preserving fine-scale spatial structure under realistic survey conditions. The developed system, named EcoVision, establishes a practical foundation for scalable, high-resolution salt marsh monitoring, demonstrating how AI-driven workflows can translate pixel-level predictions into ecologically interpretable metrics.

\end{abstract}



\begin{keyword}
{method}, {dataset}, {pipeline}, {segform}, {convnext}, {training}, {preprocessing}, {annotation}, {augment}, {collection}, {uav survey}, {data split}



\end{keyword}

\end{frontmatter}



\section{Introduction}
Among coastal landscapes, vegetation uniquely underpins ecosystem service delivery and functioning, regulating biodiversity patterns, habitat structure, and key ecological processes \cite{barbier2011value}. Accurate, high-resolution monitoring is critical for assessing environmental change, guiding conservation strategies, and informing predictive ecological models \cite{waldchen2017plant,ipbes2024global}. Despite their long-standing role in ecology, field surveys depend heavily on human judgement and manual effort, introducing variability and limiting scalability for both temporal and spatial coverage \cite{bertacchi2019using,elzinga2002monitoring}. 

These constraints are particularly pronounced in dynamic or inaccessible habitats, such as salt marshes, where interestingly species of quite different traits and morphology can co-exist in physical niches defined by mere tens of centimetres. The consequences of these very strong physical niches, yet also expansive, inaccessible, and heterogeneous settings, mean traditional surveys are both onerous, focused typically on transects, and often generalise a marsh rather than perhaps finding specific areas of concern that might be missed by linear walks \cite{Stevens2023MarshResilience}. Satellite imagery might counter this limitation; however, satellite data lacks the spatial resolution to identify species and quantify their associated ecosystem service value \cite{Reddy2021RemoteSensingBiodiversity}, and it also lacks the temporal flexibility necessary for species-level analysis \cite{yang2025unmixing,chang2025time}.
Emerging UAV-based remote sensing offers high-resolution, repeatable imaging capabilities that capture fine-scale vegetation patterns. At the same time, machine learning methods enable the automated identification, segmentation, and dominance assessment of species \cite{kattenborn2019convolutional}. By combining UAV imagery with computer vision and deep learning, the proposed framework addresses critical limitations of conventional monitoring, providing scalable, objective, and ecologically interpretable data. This integration also supports further advances such as near real-time ecosystem assessment, invasive species management, and evidence-based conservation planning, bridging the gap between high-resolution imagery and actionable ecological insight \cite{decastro2021uavs, cruz2023improving, Stevens2023MarshResilience}. 

\subsection{Research Objectives}
This research aims to develop and evaluate an integrated UAV-AI framework for automated plant species identification and dominance assessment in salt marsh ecosystems. Existing approaches largely focus on species presence/absence mapping or pixel-level classification, offering limited insight into competitive dynamics and ecological dominance within defined spatial units. EcoVision addresses this gap by combining high-resolution UAV imagery with advanced computer vision and deep learning to generate ecologically meaningful, patch-level outputs.
The following are the primary objectives of this study: 
\begin{itemize}
\item[Obj1:] to acquire high-resolution UAV imagery capable of supporting species-level analysis across heterogeneous coastal landscapes \cite{turner2016uavs, aasen2018quantitative}. 
\item[Obj2:] to implement and assess transformer-based semantic segmentation and blob-level classification models \cite{xie2021segformer} for accurate vegetation delineation and species identification.
\item[Obj3:] to design a spatial aggregation strategy that translates model predictions into quantitative dominance scores at a 2$\times$2 m patch scale as per JNCC guidelines \cite{JNCC2006NVCHandbook}. 
\item[Obj4:] to evaluate the robustness and generalisability of the proposed pipeline across varying environmental conditions and datasets, ensuring suitability for real-world ecological monitoring.
\end{itemize}

\subsection{Contributions of This Study}
Based on these objectives, the following are the contributions of this study to the field of AI-driven ecological monitoring.

\begin{itemize}
\item[C1:] An end-to-end UAV–AI pipeline that integrates high-resolution aerial data acquisition, species-level semantic segmentation, blob-based classification, and patch-level dominance scoring into a unified and reproducible workflow. To our knowledge, no existing study combines transformer-based semantic segmentation, object-level classification, and quadrat-scale dominance scoring in a single pipeline validated against field survey data for salt marsh ecosystems \cite{yang2025unmixing, moreno2025multi, morgan2022deep}.

\item[C2:] Application of transformer-based segmentation (SegFormer) and modern convolutional architectures (ConvNeXt) for fine-grained vegetation analysis in complex salt marsh environments, addressing challenges such as overlapping canopies, background noise, and seasonal variability. This contributes empirical evidence for the suitability of state-of-the-art vision models in real-world ecological settings.

\item[C3:] A 2$\times$2~m patch-based dominance framework that provides a scalable, policy-aligned method for quantifying species dominance consistent with JNCC National Vegetation Classification guidelines \cite{JNCC2006NVCHandbook}. This spatially explicit approach extends beyond presence--absence mapping to support quantitative assessment of competitive dynamics and habitat condition, directly comparable to field quadrat surveys.
\end{itemize}
 The emphasis of this research is on modularity and reproducibility \cite{Pineau2021Reproducibility}, offering a framework that can be adapted to other ecosystems \cite{decastro2021uavs}, species groups, and monitoring objectives. Collectively, these contributions advance the practical integration of Artificial Intelligence (AI) and Unmanned Aerial Vehicle (UAV) technologies into ecological monitoring, supporting more objective, scalable, and actionable environmental assessment.

\section{Related Work}
Traditional ecological surveys remain fundamental to biodiversity monitoring, yet they exhibit well‑documented limitations that constrain their effectiveness, particularly for programmes requiring consistent spatial and temporal coverage. Field-based approaches rely heavily on manual species identification and visual estimation, introducing observer bias and uncertainty - especially for cryptic, morphologically similar, or rare species \cite{waldchen2017plant,bertacchi2019using}. These surveys are also labour‑intensive and time‑consuming, often requiring specialist teams to work in remote or physically challenging environments, which limits the frequency and scale at which monitoring can realistically occur \cite{elzinga2002monitoring}.
Scalability poses a further challenge. Extending surveys across large, remote, or environmentally dynamic landscapes, such as wetlands or coastal systems, requires resources far exceeding what most conservation or land-management programmes can sustain. Yet such repeated, fine‑scale monitoring is essential for detecting seasonal or inter-annual ecological change and avoiding gaps in biodiversity records \cite{moreno2025multi}. While current field methods may be optimised in terms of cost–benefit, a substantial spatial–temporal gap remains between what ecologists can practically survey and the resolution of data required for timely, actionable management.
Another limitation lies in the delay between data collection and ecological interpretation. Traditional surveys require significant expert verification and taxonomic quality control, which extends the time before results can be incorporated into decision-making. This latency reduces the utility of field data for adaptive management, especially following rapid or transient disturbances such as storm events or short-lived but ecologically consequential anthropogenic impacts \cite{moreno2025multi}.
Taken together, these constraints, i.e., observer subjectivity, limited coverage, intensive labour requirements, and delays in data availability, highlight the need to augment (not replace) conventional survey methods with complementary, technology‑driven tools. UAV imaging and machine learning offer a promising addition to the land‑management “toolbox”, providing rapid, consistent, and objective assessments that can be later supported or quality‑checked by ecologists with far less time and cost\cite{vanalphen2024uav, gray2022drones}. Recent work demonstrates that convolutional neural networks can accurately segment and identify plant species from high‑resolution imagery \cite{kattenborn2019convolutional}, enabling fast, good‑faith estimates that ecologists can validate more efficiently than by conducting full field surveys.

\subsection{Role of AI in Environmental Monitoring}
Deep learning architectures marked a significant advancement. \cite{GonzalezPerez2022} demonstrated the superiority of convolutional neural networks (CNNs) like U-Net and DeepLabv3 over traditional ML (SVM/RF) for classifying 17 coastal wetland classes, including salt marsh vegetation, using UAS multispectral imagery and LiDAR-derived CHMs. U-Net achieved overall accuracies of 83.8–85.3\%, outperforming ML models (57–71\%), with visible-band imagery proving nearly as effective as multispectral when paired with DL. Subsequent studies confirmed this trend. \cite{Bai2025} applied U-Net and SegNet to UAV multispectral imagery in the Yellow River Delta, attaining 94\%+ accuracy for typical salt marsh vegetation types, far exceeding RF and XGBoost.
Multi-sensor fusion and habitat-specific modeling have further enhanced precision. \cite{Moreno2025} fused RGB, MS, and HS UAV data with ML algorithms (RF, SVM, Spectral Angle Mapper), achieving 97\% overall accuracy for species classification in Irish salt marshes, surpassing single-sensor results (max 92\% with HS). Band stacking and PCA dimensionality reduction effectively handled spectral overlap. \cite{Agate2026} used RF on UAV multispectral imagery, DSMs, and textural features to map nine National Vegetation Classification communities in UK restored and established salt marshes, yielding a mean accuracy of 94.7\% across sites of varying restoration ages. DSM proved the most important predictor due to elevation-tidal inundation relationships.
Biomass estimation, critical for carbon accounting, has also benefited. \cite{Liu2025} and \cite{Curcio2024} integrated UAV MS and LiDAR with ML for species-level above-ground biomass (AGB) modeling, achieving up to 99\% precision through habitat-specific approaches that account for seasonal and stress-related variations (e.g., spring salinity impacts). Deep learning semantic segmentation has additionally enabled invasive species mapping, such as Spartina anglica, with high mIOU scores.
These studies highlight key trends: DL/CV models consistently outperform traditional ML in heterogeneous landscapes; multi-sensor fusion and structural data (LiDAR/DSM) improve boundary delineation and accuracy; and UAV-AI workflows support restoration monitoring, biodiversity assessment, and climate resilience tracking without extensive fieldwork. However, labelled data scarcity, computational demands, seasonal variability, and upscaling are still major issues. Therefore, application of Transformer-based models, self-supervised learning, real-time edge AI on UAVs, and multi-temporal integration for dynamic ecosystem modelling ahould be the next step forward.

\subsection{UAV-Based Ecological and Environmental Monitoring}
The integration of UAVs with AI and computer vision has significantly advanced ecological monitoring by addressing key limitations of traditional field surveys. Operating at low altitudes, UAVs provide centimetre-scale, high-resolution imagery across heterogeneous and often inaccessible coastal environments such as salt marshes, wetlands, and intertidal zones \cite{vanalphen2024uav, moreno2025multi}. Their capacity for repeated flights over fixed plots enables consistent temporal monitoring, facilitating the detection of short-term and seasonal vegetation changes that are difficult to quantify using conventional methods \cite{yang2025unmixing, pottker2023convolutional}.
When combined with AI-driven analytical pipelines, UAV imagery supports automated species detection, pixel-level semantic segmentation, and patch-scale estimation of vegetation dominance and biomass \cite{pottker2023convolutional, gao2026high}. Advanced models, including convolutional neural networks (CNNs) and transformer-based segmenters, reduce subjectivity, enhance repeatability, and allow scalable analysis of large datasets \cite{decastro2021uavs}. These capabilities generate ecologically meaningful outputs such as species distribution maps, dominance layers, and habitat descriptors that directly support evidence-based conservation planning and land management \cite{gautam2025detection, morgan2022deep}.
UAVs have proven particularly valuable in dynamic ecosystems like salt marshes, where vegetation undergoes rapid shifts due to tidal cycles, salinity stress, and restoration activities. Early studies demonstrated their ability to map vegetation at species-level resolution more efficiently than manual surveys \cite{bertacchi2019using}. As UAV applications expanded across coastal, agricultural, and forestry ecosystems, integration with machine learning further improved detection and classification of invasive, structurally similar, or spatially intermixed species \cite{bertacchi2019using, cruz2023improving}. Meta-analyses consistently highlight superior spatial detail and classification accuracy compared to traditional surveys, although performance remains sensitive to sensor selection, flight parameters, and species complexity \cite{deng2022comparison, waldchen2017plant}.
Nevertheless, challenges persist in transforming raw UAV imagery into standardised, ecologically interpretable products and ensuring workflow consistency across different sites and species. These limitations highlight the need for integrated end-to-end frameworks that unify sensing, segmentation, classification, and ecological inference. Initiatives such as EcoVision represent important steps toward developing such unified operational pipelines, effectively bridging high-resolution environmental imagery with actionable ecological indicators.
Overall, the synergy between UAVs and AI provides a powerful complement to traditional monitoring methods, delivering consistent, spatially explicit assessments that enhance precision, support restoration evaluation, and strengthen conservation outcomes in vulnerable coastal ecosystems.

\subsection{Computer Vision for Plant and Vegetation Analysis}
Computer vision has become central to vegetation analysis, allowing automated extraction of structure, composition and species-level information from imagery. Early feature‑based approaches were constrained by lighting variability, background clutter, and species similarity \cite{waldchen2017plant}. Deep learning, especially CNNs, overcame many of these issues by learning hierarchical representations that capture texture, morphology, and contextual cues critical for differentiating plant species \cite{milioto2018real,morgan2022deep}.

Advanced semantic and instance segmentation models now allow pixel or object‑level classification with high spatial precision, improving the delineation of overlapping plants and complex canopy structures \cite{bertacchi2019using}. Transformer based architectures further enhance performance by leveraging long-range spatial dependencies, which is particularly valuable in dense marsh environments.
Despite these improvements, many studies stop at the segmentation stage and do not translate outputs into ecological metrics such as species dominance, patch structure, or vegetation dynamics \cite{kattenborn2019convolutional}. Deep learning now dominates species classification and vegetation segmentation due to its capacity to learn discriminative features from complex imagery. CNN architectures such as ResNet, EfficientNet, and ConvNeXt have demonstrated strong performance when combined with high-resolution UAV data, enabling fine‑grained species identification \cite{liu2022convnet, pottker2023convolutional}.

Semantic segmentation models, including U-Net variants, DeepLabV3+, and transformer-based models like SegFormer, provide robust pixel-level predictions even in challenging environments with complex illumination and vegetation structure \cite{morgan2022deep}. Attention mechanisms and multi-scale feature extraction further enhance boundary accuracy and class separation.
Persistent limitations, however, include generalisation across sites, the scarcity of annotated ecological datasets, and high computational demands for large-area imagery \cite{kattenborn2019convolutional}. Hybrid pipelines that decouple segmentation and classification increasingly show promise for improving accuracy and interpretability, aligning with EcoVision’s modular architecture \cite{gautam2025detection}.

\subsection{Gaps in Existing Approaches}
Despite significant advances, key gaps remain in UAV‑ and deep learning–based vegetation analysis. Many studies focus narrowly on either segmentation or classification rather than integrating these tasks into pipelines that reflect real ecological workflows. As a result, outputs often lack the ecological structure needed to inform habitat assessments, such as species dominance, patch organisation, or competitive dynamics \cite{deng2022comparison}.
Generalisation remains another major challenge. Most models are trained on limited datasets that fail to represent the variation introduced by differing sensors, flight conditions, seasons, or background vegetation, resulting in domain shift when methods are applied to new sites \cite{kattenborn2019convolutional}. Ecological interpretability is further constrained by a lack of uncertainty quantification and by the absence of mechanisms to convert pixel-level data into meaningful ecological units \cite{yang2025unmixing}.
Finally, computational overhead limits the scalability of existing methods, restricting their applicability for near–real-time ecological decision‑making. 

These gaps motivate the need for end‑to‑end, ecologically grounded systems such as EcoVision, that integrate sensing, segmentation, classification, and ecological interpretation in a manner that balances accuracy, interpretability, and operational feasibility.
EcoVision directly addresses these gaps by coupling SegFormer-based segmentation with ConvNeXt classification and quadrat-scale dominance scoring in a single, field-validated pipeline \cite{xie2021segformer, liu2022convnet, kent2012vegetation}.

\section{Methodology}

\begin{figure}[H]
    \centering
    \
    \includegraphics[width=\linewidth]{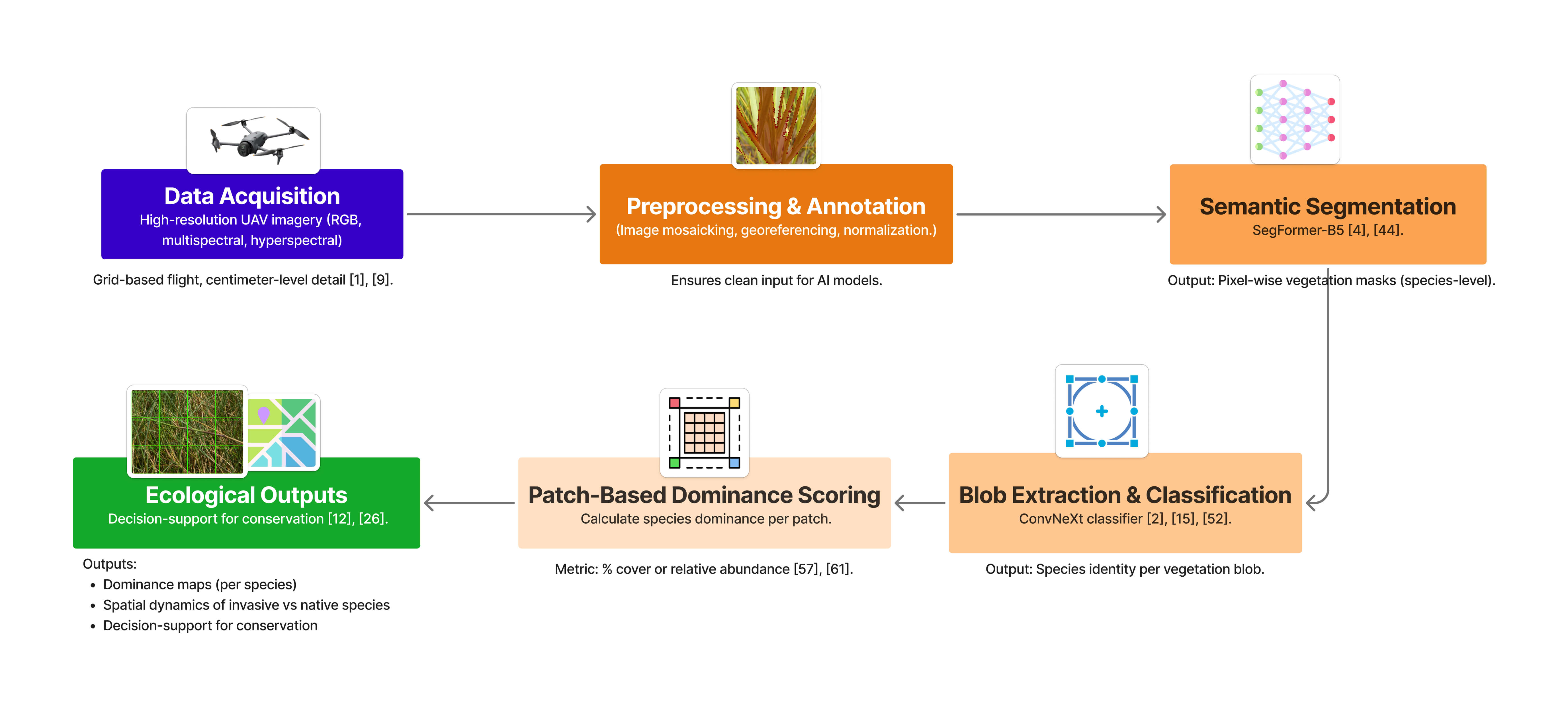}
    \caption{End-to-end flow of the EcoVision pipeline.}
    \label{fig:pipeline}
\end{figure}

EcoVision is positioned at the intersection of UAV-based ecological monitoring, modern computer vision, and ecologically interpretable machine learning. 
EcoVision adopts an end-to-end pipeline aligned with real ecological workflows \cite{decastro2021uavs} as illustrated in Figure \ref{fig:pipeline}. High-resolution UAV imagery with a ground sampling of 0.5–1 cm/pixel serves as the primary data source, while deep learning models are used not only for pixel-level accuracy but as building blocks for higher-level ecological inference \cite{moreno2025multi, ma2019deep}. 
Semantic segmentation is used to delineate vegetation extents, followed by object-level extraction and species classification, enabling the transition from raw imagery to quantified species presence. Crucially, predictions are aggregated into fixed spatial units, allowing species dominance and distribution to be assessed in a manner consistent with field-based ecological methods \cite{elzinga2002monitoring}.
EcoVision also addresses key weaknesses in existing approaches by emphasising modularity, scalability, and interpretability. The system is designed to support adaptation to new species and survey sites with minimal pipeline reconfiguration. By grounding model outputs in spatial and ecological context, EcoVision bridges the gap between computer vision research and applied environmental monitoring, positioning it as a practical and extensible framework rather than a task-specific experimental study.

\subsection{EcoVision Architecture} \label{sec:arch}
\begin{figure}[H]
    \centering
    \
    \includegraphics[width=0.8\linewidth]{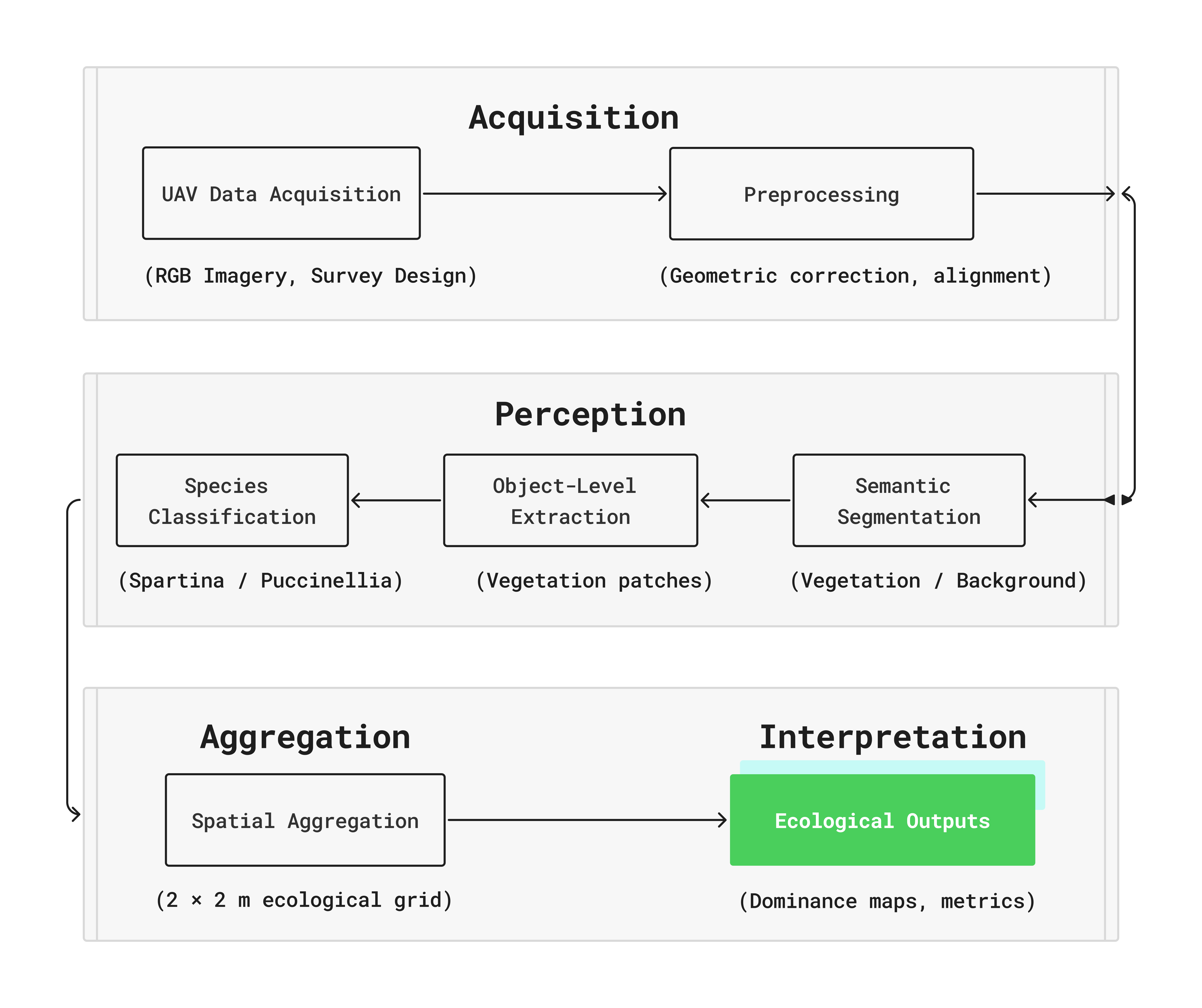}
    \caption{Conceptual architecture of the EcoVision framework. The system is organised as a modular, end-to-end pipeline. High-resolution UAV imagery is progressively transformed through perception, object-level interpretation, and spatial aggregation into ecologically meaningful outputs such as species presence and dominance.}
    \label{fig:architecture}
\end{figure}
Figure \ref{fig:architecture} shows the complete architecture of EcoVision, consisting of three layers. 
The \textbf{acquisition} layer explains how to preprocess the data captured using UAV camera, this include geometric correlation, alignment and improving image quality. 
At the \textbf{perception} layer, semantic segmentation models identify vegetation and non-vegetation regions at the pixel level, establishing spatial structure within each scene \cite{xie2021segformer,long2015fully}. This is followed by object-level extraction, where contiguous vegetation regions are isolated to enable species-level reasoning \cite{haralick1992computer, gonzalez2018digital}. A dedicated classification module assigns species labels to these regions, allowing heterogeneous vegetation to be decomposed into ecologically relevant units (\textbf{interpretation}) \cite{liu2022convnet}.
Aggregation involves predictions within fixed 2$\times$2m spatial grids, translating model outputs into species dominance scores computed as proportional areal coverage per grid cell, a unit consistent with conventional field-based ecological sampling frameworks \cite{kent2012vegetation}.
By separating acquisition, perception, interpretation, and aggregation into distinct but connected stages as described in Figure~\ref{fig:architecture}, EcoVision emphasises clarity, adaptability, and ecological validity, enabling robust monitoring across sites and time without redesigning the entire system \cite{decastro2021uavs}. Following is the description of the steps involved:

\begin{itemize}
\item High-resolution UAV imagery is first acquired under standardised survey conditions to ensure consistency across flights and sites \cite{turner2016uavs, anderson2013lightweight}. These images form the sole input to the computational pipeline. 
\item SegFormer-B5, a transformer-based semantic segmentation model, is then applied to isolate vegetation from background elements such as water, sediment, and exposed substrate, producing structured vegetation masks that capture spatial extent and patch boundaries \cite{xie2021segformer}.
\item Connected component analysis is used to extract contiguous vegetation regions from these masks, transforming pixel-level predictions into discrete vegetation units, referred to hereafter as \textbf{blobs}, that serve as the fundamental objects for downstream classification \cite{haralick1992computer, gonzalez2018digital}. 
\item Blob regions are subsequently classified using a ConvNext convolutional neural network to assign species-level labels, enabling discrimination between dominant marsh taxa and non-target classes \cite{liu2022convnet}.  Only classifications exceeding a defined confidence threshold are forwarded to the aggregation stage. Classified vegetation units are then aggregated within fixed 2 $\times$ 2 m grids (see Section~\ref{sec:arch})
\item Species dominance is computed as proportional areal coverage per grid cell (Equation~\ref{eq:dominance}), aligning model outputs with established ecological sampling scales. 
\item Final outputs are visualised as georeferenced maps and a dominance heatmap, supporting spatial analysis, long-term monitoring, and conservation management \cite{elzinga2002monitoring, Stevens2023MarshResilience}. 
\end{itemize}

\subsubsection{Species of Interest }
\begin{figure}[H]
    \centering
    \begin{subfigure}{.5\textwidth}
       \centering
       \includegraphics[width=1\linewidth]{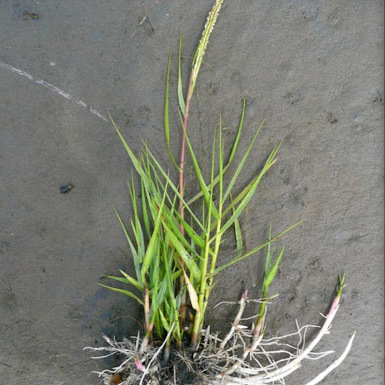}
       \caption{Spartina maritima}
       \label{fig:sub1}
        
    \end{subfigure}%
    \begin{subfigure}{.5\textwidth}
       \centering
       \includegraphics[width=1\linewidth]{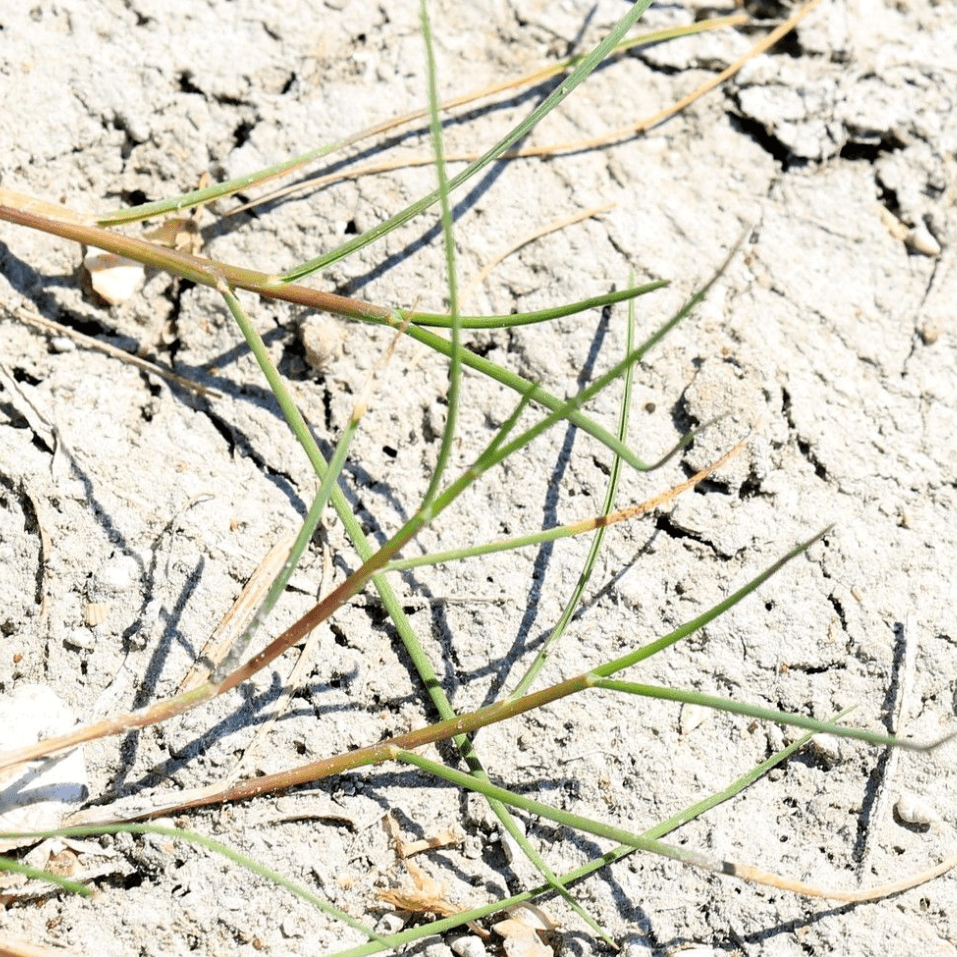}
       \caption{Puccinellia maritima}
       \label{fig:sub2}
        
    \end{subfigure}
\caption{Representative field photographs of the two target species. (a) Spartina maritima: dense, erect culms with narrow leaves characteristic of pioneer low-marsh zones. (b) Puccinellia maritima: prostrate tussock growth form typical of mid-marsh communities. These contrasting canopy structures underpin their distinguishability under UAV-based imaging.}
    \label{fig:species}
\end{figure}
This study focused on two dominant British salt marsh species (see Figure~\ref{fig:species}): \textit{Spartina maritima} and \textit{Puccinellia maritima}. \textit{Spartina maritima} is a native pioneer cordgrass occupying low-marsh and pioneer zones, characterised by dense, erect culms and narrow leaves adapted to frequent tidal inundation \cite{adam1990saltmarsh, bertness1991zonation}. \textit{Puccinellia maritima} is a perennial saltmarsh grass forming prostrate tussock communities in mid-marsh zones, occupying higher intertidal elevations with reduced submergence frequency \cite{adam1990saltmarsh, JNCC2006NVCHandbook}. Together, these species represent the dominant vegetated zones of British Atlantic saltmarshes and constitute key indicators of habitat condition and succession \cite{kent2012vegetation, barbier2011value}. Their contrasting growth forms, erect and dense versus prostrate and tussock-forming, produce distinct canopy architectures detectable under high-resolution UAV imaging, making them suitable targets for deep learning-based segmentation and classification \cite{bertacchi2019using, moreno2025multi, yang2025unmixing}. Additional non-target classes, including bare sediment, tidal water, and mixed background vegetation, were incorporated into the classification framework to reduce false positives and improve discrimination in heterogeneous marsh mosaics \cite{kattenborn2019convolutional}.

\subsection{Study Area}\label{sec:study-area}
\begin{figure}[H]
    \centering
    \begin{subfigure}{.48\textwidth}
\includegraphics[width=\linewidth]{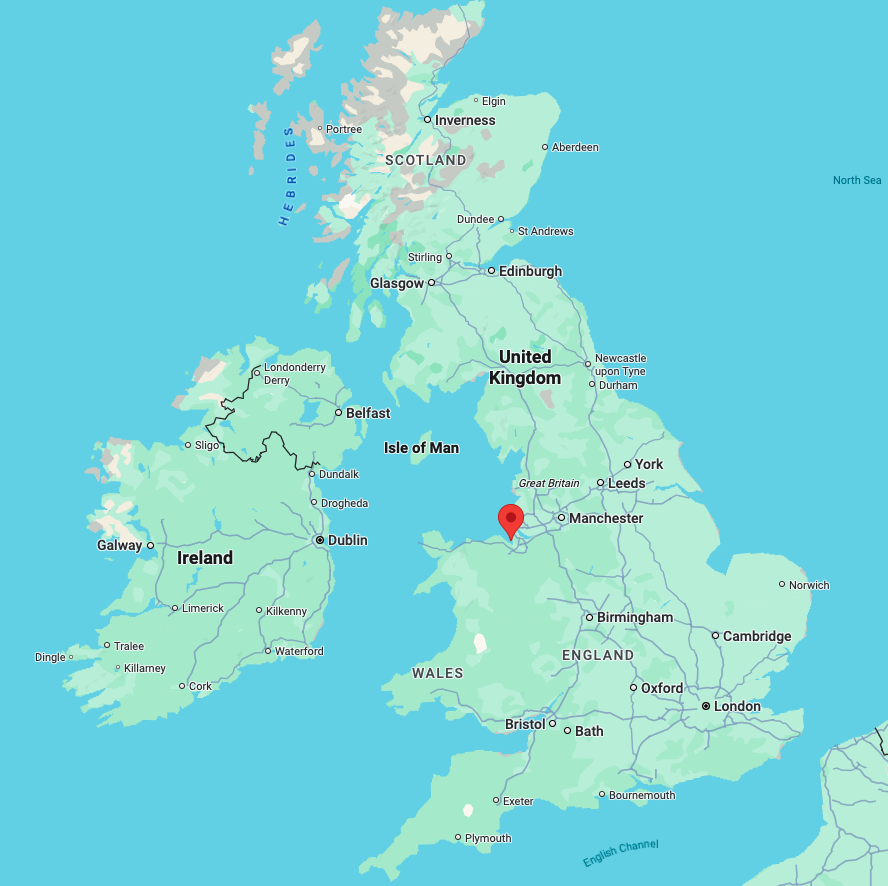}
        \caption{A map showing the site location} 
        \label{fig:wales-uk}
    \end{subfigure}
    \hfill
    \begin{subfigure}{.48\textwidth}
\includegraphics[width=\linewidth]{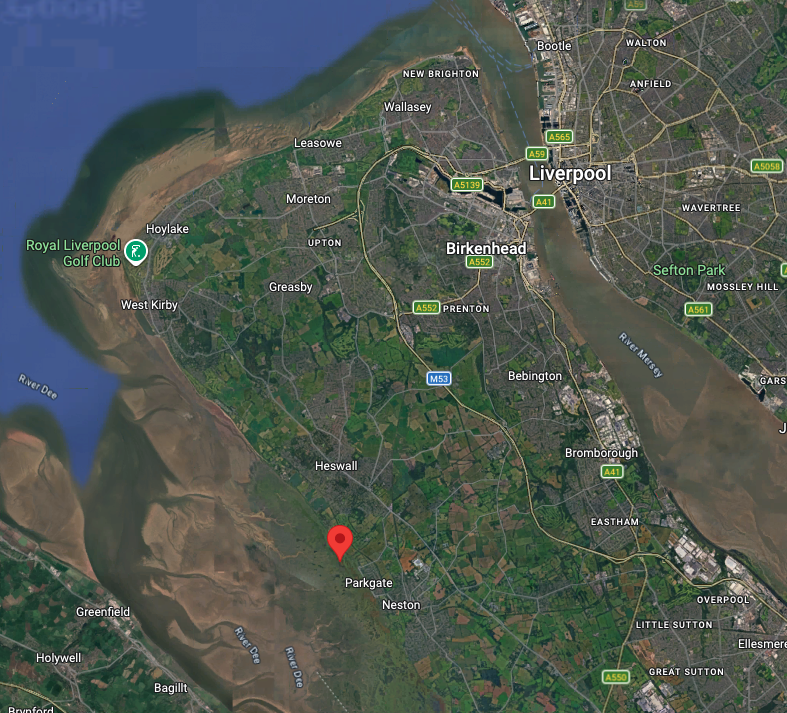}
        \caption{Aerial view showing the survey quadrat locations} 
        \label{fig:closeup} 
    \end{subfigure}
      
    \caption{Study area location. (a) Regional context showing the survey site on the North Wales coast at the mouth of the River Clwyd, Rhyl, Denbighshire (53°18'N, 3°06'W). (b) Detailed view of the Foryd Bay intertidal salt marsh, showing the heterogeneous vegetation mosaic of \textit{Spartina maritima} and \textit{Puccinellia maritima} communities across pioneer, low-marsh, and mid-marsh zones surveyed by UAV}
    \label{fig:study-area}
\end{figure}

UAV surveys were conducted at the Foryd Bay intertidal salt marsh, located at the mouth of the River Clwyd, Rhyl, Denbighshire, North Wales, UK (53°18'1.85"N, 3°05'59.38"W; Figure~\ref{fig:study-area}). The site lies within the transition zone between the River Clwyd estuary and the Irish Sea, forming a representative example of a British Atlantic intertidal salt marsh subject to regular tidal inundation and characterised by halophytic vegetation 
communities \cite{adam1990saltmarsh, JNCC2006NVCHandbook}.
The survey area encompasses pioneer, low-marsh, and mid-marsh zones, supporting mixed communities dominated by \textit{Spartina maritima} and \textit{Puccinellia maritima}, with additional background classes including bare sediment, tidal channels, and mixed vegetation. This heterogeneous mosaic, combined with the site's accessibility at low tide and its representativeness of 
British Atlantic salt marsh habitats made it well-suited for evaluating the EcoVision pipeline under real-world monitoring conditions \cite{Stevens2023MarshResilience, 
bertacchi2019using}.

\subsection{Data Acquisition} \label{sec:data-acq}
UAV data were collected using a DJI Mavic 4 Pro equipped with a 20-megapixel, 1-inch CMOS RGB sensor and a 24 mm equivalent lens (see Table~\ref{tab:uav_params}). This configuration provides a practical balance between spatial resolution, flight stability, and operational accessibility, without reliance on specialised multispectral payloads,  consistent with recent demonstrations of RGB sufficiency for species-level wetland vegetation mapping \cite{bertacchi2019using}. The sensor provides sufficient spatial detail to resolve fine-scale vegetation morphology required for species-level segmentation and classification \cite{aasen2018quantitative}. RGB imagery was captured in both RAW and JPEG formats, preserving spectral fidelity for analysis while maintaining efficient data handling for large survey volumes. Flights were executed at approximately 10~m above ground level (AGL), yielding a ground sampling distance (GSD) of 0.5–1.0~cm per pixel. A grid-based survey pattern was adopted with 70–80\% forward and lateral image overlap to ensure robust photogrammetric reconstruction \cite{iglhaut2019structure, colomina2014unmanned}. Survey coverage targeted representative salt marsh zones, including pioneer, low-marsh, and mid-marsh areas, aligned with established ecological zonation frameworks \cite{adam1990saltmarsh}. 

\begin{table}[H]
    \centering
\caption{UAV flight and sensor parameters used for data acquisition.}
\label{tab:uav_params} 
    \begin{tabular}{ll}\toprule
         \textbf{Parameter}& \textbf{Value}\\ \midrule
 Platform&DJI Mavic 4 Pro \\
 Sensor&1-inch CMOS RGB, 20 MP \\
 Focal length&24 mm equivalent\\
 Altitude (AGL)&$\sim{10 m}$\\
 GSD& 0.5–1.0 cm/pixel\\
 Image overlap&70–80\% (forward and lateral)\\
 Image formats&RAW and JPEG\\
 Survey zones& Pioneer, low-marsh, mid-marsh \\
 Tide condition&Low tide\\ \bottomrule
    \end{tabular}

\end{table}

All UAV flights were conducted during low-tide windows to maximise visibility of vegetated patches and minimise tidal interference \cite{turner2016uavs,vanalphen2024uav}. Data collection was restricted to stable atmospheric conditions, with clear or uniform overcast skies preferred to reduce shadowing and illumination variability \cite{aasen2018quantitative}. Imagery was geotagged using onboard GPS and subsequently refined during photogrammetric processing to improve spatial accuracy \cite{iglhaut2019structure}. Captured datasets were reserved exclusively for validation and analysis, ensuring independence from training data.

Model training data were sourced from global biodiversity repositories, including iNaturalist \cite{inaturalist2023} and GBIF \cite{gbif2023}, using programmatic API access and species-specific filtering \cite{gbif2025}. Manual quality control was applied to exclude blurred, dried, or off-target specimens, retaining approximately 240 and 220 for each species ($\approx$460 images total across both repositories). These images captured variation in environmental conditions, growth stages, and viewing angles, enhancing model generalisation across real-world survey conditions. UAV imagery was stored separately as independent validation data. All datasets followed consistent naming conventions and metadata standards to support traceability, reproducibility, and seamless integration with segmentation and classification pipelines \cite{Pineau2021Reproducibility}. This two-tiered data strategy combined taxonomically diverse public imagery with site-specific UAV data used exclusively for validation, balancing taxonomic breadth with ecological relevance \cite{moreno2025multi}.

\subsection{Dataset Preparation and Annotation}
Building on the two-tiered data strategy described in Section~\ref{sec:data-acq}, imagery from both sources underwent systematic preparation to construct a robust dataset for deep learning. UAV data preserved morphological detail at 0.5–1.0~cm resolution, while the public repository images \cite{inaturalist2023, gbif2025} provided visual variability essential for model generalisability \cite{waldchen2017plant, ma2019deep}.

\subsubsection{Annotation Strategy}
Species-level annotations were produced manually using polygonal delineation in LabelMe \cite{russell2007labelme}. Masks captured precise vegetation boundaries, as pixel-accurate segmentation is required to resolve overlapping canopies in salt marsh environments \cite{kattenborn2019convolutional}. Annotations were stored in JSON format and converted to binary and multi-class PNG masks via a custom Python script. Each image-mask pair adhered to strict naming conventions to support traceability and seamless integration with the segmentation and classification pipeline \cite{Pineau2021Reproducibility}.

\subsubsection{Preprocessing and Augmentation}
All images were resized to 512 $\times$ 512 pixels to standardise spatial input for deep learning models, consistent with the Segformer-B5 input specification. Preprocessing prioritised preservation of vegetation structure, applying minimal normalisation to avoid suppressing ecologically relevant spectral variation. To mitigate class imbalance and improve generalisation, a probabilistic augmentation pipeline was implemented using the Albumentation library \cite{buslaev2020albumentation,shorten2019survey}, including:

\begin{itemize}
    \item Horizontal flips ($p=0.5$)
    \item Random 90° rotations ($p = 0.5$) 
    \item Brightness and contrast variation ($p = 0.3$)
    \item Affine transformations: translation, scaling, rotation within $\pm20°$ ($p = 0.5$) 
    \item Gaussian noise injection ($p = 0.2$)
\end{itemize}

Each original image–mask pair generated five augmented variants, producing approximately \textbf{~2,300} augmented samples in total and substantially strengthening the dataset for training deep neural networks \cite{shorten2019survey}.

\subsubsection{Data Quality Control and Validation}
All images underwent manual quality screening to exclude blurred captures, dried or degraded specimens, and ecologically irrelevant samples following criteria established for plant identification datasets \cite{waldchen2017plant}. The final dataset was partitioned into training (70\%), validation (15\%), and testing (15\%) subsets using stratified random sampling to preserve class distribution across splits \cite{Pineau2021Reproducibility}. A fixed random seed was applied to ensure reproducibility of all reported results. UAV imagery was reserved exclusively for independent testing, as established in Section~\ref{sec:data-acq} \cite{ma2019deep}. This deliberate domain separation addresses a known challenge in ecological deep learning, where models trained on close-range or herbarium images frequently underperform on UAV-acquired field imagery \cite{ waldchen2017plant}.
\begin{table}[H]
\centering
\caption{Dataset composition before and after augmentation.}
\label{tab:dataset}
\begin{tabular}{>{\raggedright\arraybackslash}p{0.12\linewidth}>{\raggedright\arraybackslash}p{0.15\linewidth}>{\centering\arraybackslash}p{0.13\linewidth}>{\centering\arraybackslash}p{0.22\linewidth}>{\centering\arraybackslash}p{0.22\linewidth}}\toprule

\textbf{Source} & \textbf{Class} & \textbf{Raw images} & 
\textbf{After Manual QC}& \textbf{After Augmentation}\\\midrule

iNaturalist & \textit{Spartina maritima} & ~209& ~80& ~400\\
GBIF & \textit{Spartina maritima} & ~364& ~160& ~800\\
iNaturalist & \textit{Puccinellia maritima} & ~146& ~78& ~390\\
GBIF & \textit{Puccinellia maritima} & ~293& ~142& ~710\\

\textbf{Total} & & \textbf{~1,012}& \textbf{~460}& \textbf{~2,300}\\ \bottomrule

\end{tabular}
\end{table}

\subsection{Data processing and classification}
EcoVision addresses vegetation mapping as a hierarchical perception problem. At the pixel level, the task is formulated as multi-class semantic segmentation, separating background, \textit{Spartina maritima}, and \textit{Puccinellia maritima}. At the object level, contiguous vegetation patches are treated as individual units for species classification. Finally, dominance estimation is framed as a spatial aggregation problem, where species-specific cover is quantified within ecologically meaningful 2$\times$2m grid cells, a scale consistent with standard quadrat-based field survey protocols \cite{kent2012vegetation, elzinga2002monitoring}. This formulation ensures consistency between computer vision outputs and traditional ecological survey metrics \cite{decastro2021uavs}.

Given a patch $P$ comprising $N$ target species, the dominance score of species $s$ is calculated as:

\begin{equation} D_s(P) = \frac{A_s(P)}{\displaystyle\sum_{i=1}^{N} A_i(P)} \times 100 \label{eq:dominance} \end{equation}
Where:

\noindent \textbf{$A_s(P)$} =  the total area (in pixels or m$^2$) covered by species $s$ within patch $P$ \\
\textbf{$N$} =  the total number of target species ($N = 2$: \textit{Spartina maritima} and \textit{Puccinellia maritima})
 \\
 \textbf{$D_s(P)$} = the resulting dominance percentage of species $s$ in patch $P$\\

Background pixels are excluded from the denominator, ensuring dominance is expressed as a proportion of total \emph{vegetated} area within each grid cell. This yields a standardised proportional cover metric consistent with ecological survey conventions \cite{kent2012vegetation}.

\subsubsection{Segmentation Models}
\begin{figure}[H]
    \centering
    \includegraphics[width=0.75\linewidth]{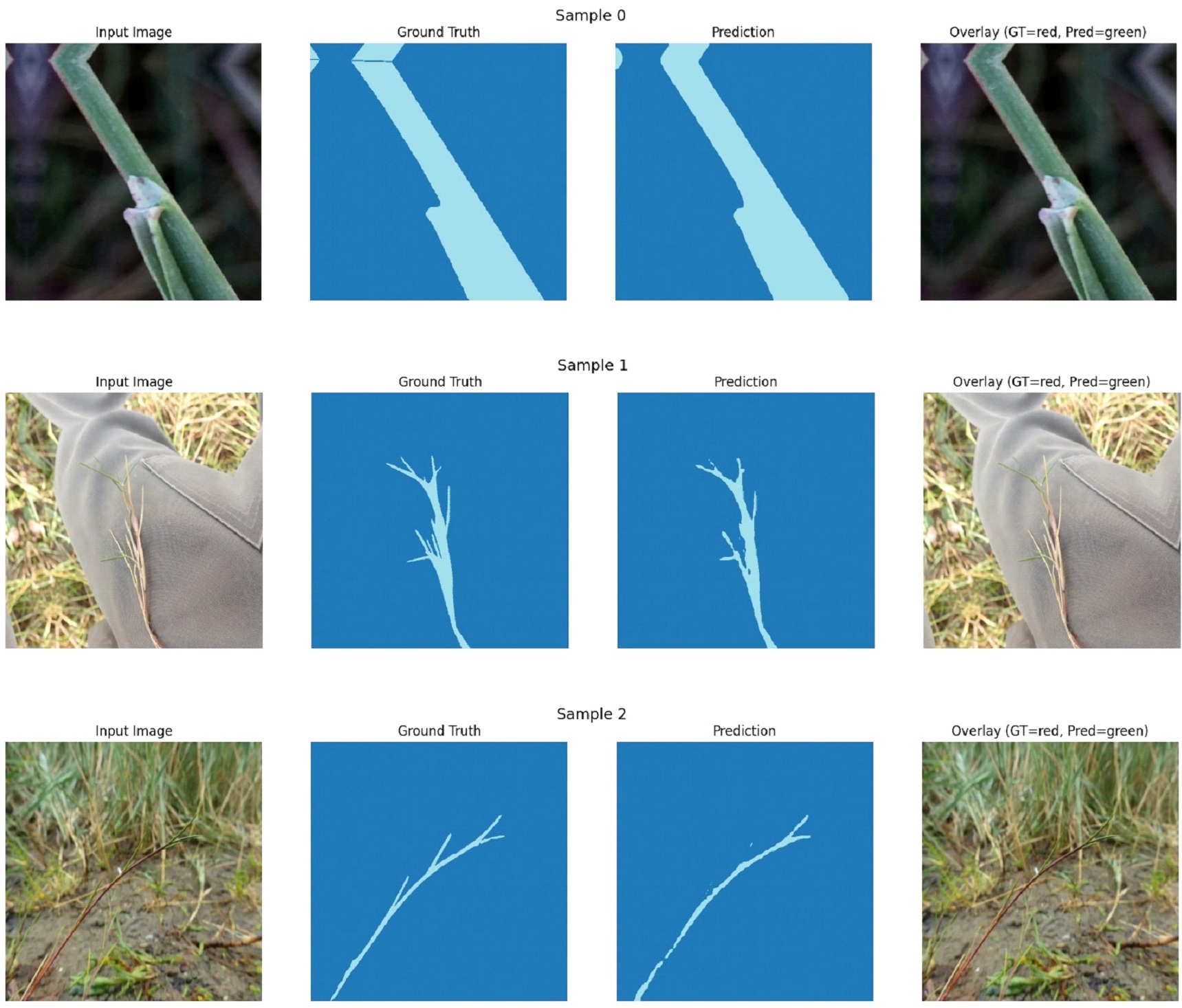}
    \caption{SegFormer-B5 architecture used for semantic segmentation of UAV salt marsh imagery. The hierarchical Mix Transformer (MiT-B5) encoder extracts multi-scale features at four resolution stages, while the lightweight MLP decoder fuses them into a pixel-level prediction map producing three semantic classes: background, \textit{Spartina maritima}, and \textit{Puccinellia maritima} \cite{xie2021segformer}.}
    \label{fig:segmentation-overlay}
\end{figure}
Semantic segmentation provides the spatial foundation of the pipeline by delineating vegetation boundaries and suppressing non-vegetated background elements. Transformer-based architectures were selected to capture both local texture and global spatial context inherent in heterogeneous salt marsh environments, effectively overcoming the receptive field limitations of traditional convolutional approaches \cite{xie2021segformer,moreno2025multi}. \\
SegFormer-B5 uses a hierarchical transformer (MiT-B5) encoder with a lightweight MLP decoder, enabling high-resolution segmentation without explicit positional encoding, a design property particularly suitable for fine-grained vegetation boundary detection \cite{xie2021segformer}. The model was initialised with ImageNet-pretrained weights via the HuggingFace Transformers library \cite{huggingface_transformers_2023}, and fine-tuned on 512 $\times$ 512 UAV imagery. Three semantic classes were defined: background, \textit{Spartina maritima}, and \textit{Puccinellia maritima}. Cross-entropy loss with class weighting was applied to mitigate the influence of class imbalance \cite{long2015fully}. This architecture was selected for its ability to preserve fine vegetation boundaries while maintaining robustness under variable illumination and canopy structure \cite{ma2019deep}.

\subsubsection{Classification Model}
Species-level discrimination was performed using ConvNeXt, a convolutional network that incorporates design elements inspired by Vision Transformers while maintaining CNN computational efficiency and hierarchical feature extraction for fine-grained classification \cite{liu2022convnet}. Cropped vegetation blob patches (224$\times$224 pixel) were classified into \textit{Spartina maritima} and \textit{Puccinellia maritima} classes. ImageNet pretrained weights were fine-tuned to adapt feature representations to salt marsh canopy morphology, a strategy particularly effective under limited labelled data conditions \cite{ma2019deep, kattenborn2019convolutional, pottker2023convolutional}. The classifier head was replaced with a two-class linear output layer, and the full backbone was jointly fine-tuned with the classification head. 

\begin{table}[H]
\centering
\caption{Training configuration for SegFormer-B5 and ConvNeXt-Base.}
\label{tab:training_config}
\begin{tabular}{lll}
\hline
\textbf{Parameter} & \textbf{SegFormer-B5} & 
\textbf{ConvNeXt-Base} \\
\hline
Task & Semantic segmentation & Species classification \\
Input size & 512$\times$512 px & 224$\times$224 px \\
Pretrained weights & ImageNet (ADE20K) & ImageNet \\
Batch size & 2 & 32 \\
Epochs & 20 & 15\\
Learning rate & $1\times10^{-4}$ & $1\times10^{-4}$ \\
Weight decay & $1\times10^{-5}$ & $1\times10^{-5}$ \\
Optimiser & AdamW & AdamW \\
Scheduler & Linear warm-up & Linear warm-up \\
Loss function & CrossEntropy (weighted) & CrossEntropy \\
Mixed precision & FP16 & FP16 \\
Checkpoint selection & Best val. mIoU & Best val. F1 \\
Framework & PyTorch + HuggingFace & PyTorch \\
\hline
\end{tabular}
\end{table}

\subsubsection{Loss Functions and Optimisation} \label{sec:loss}
CrossEntropyLoss was used for both segmentation and classification tasks. For segmentation, inverse-frequency class weighting was applied to mitigate the imbalance between dominant vegetation and background pixels, with background pixels set as the ignore index to exclude them from gradient computation \cite{long2015fully,lei2024deep}. The weighted cross-entropy loss is defined as:
\begin{equation}
  \mathcal{L}_{\mathrm{WCE}} = -\sum_{c=1}^{C} w_c \, y_c \log \hat{p}_c
  \label{eq:wce}
\end{equation}

\noindent where:
$C = 3$, the number of classes
        (\textit{Spartina maritima}, \textit{Puccinellia maritima}, background),\\
$y_c \in \{0,1\}$ = ground-truth one-hot label for class $c$,\\
$\hat{p}_c$ = predicted softmax probability for class $c$; \\
$w_c = \tfrac{1}{f_c} \big/ \sum_{c'} \tfrac{1}{f_{c'}}$ the inverse-frequency class weight, with $f_c$ denoting the pixel
        frequency of class $c$ in the training set. \\
 ${c'} =$ normalization loop to sum across all classes \\

This formulation \ref{eq:wce} assigns higher loss penalties to minority classes in this case \textit{Puccinellia maritima}, ensuring that their under-representation does not cause the model to default to dominant class predictions. For classification, standard unweighted cross-entropy was applied, as blob-level class distributions were more balanced following augmentation. \\\\
AdamW optimisation was used for both models, providing decoupled weight decay that improves generalisation across heterogeneous marsh conditions compared to standard stochastic gradient descent  \cite{loshchilov2017decoupled}. Mixed precision (FP16) training was applied to accelerate convergence and reduce GPU memory overhead \cite{micikevicius2017mixed}. A linear learning-rate warm-up schedule was used to stabilise optimisation during early training epochs for both models.

\subsubsection{Inference and Post-Processing}
During inference, species-specific binary masks were generated by thresholding class predictions from the segmentation model, assigning foreground values to target species pixels and background values to all other classes. Vegetation blobs were then extracted from these masks using connected component analysis 
\cite{haralick1992computer, gonzalez2018digital}, isolating contiguous vegetation patches as discrete spatial units. Small artefacts and edge fragments were removed by area thresholding, excluding blobs smaller than 100 pixels, an empirically defined threshold selected to balance noise suppression with retention of ecologically relevant patches.

Each retained blob was cropped from the original UAV imagery and classified using the ConvNeXt-Base model. Blob areas, initially measured in pixels, were converted to square metres using the known ground sampling distance (GSD = 0.5–1.0~cm/pixel) to enable ecologically interpretable area measurements 
\cite{aasen2018quantitative}. Classified blobs were spatially assigned to their corresponding grid cell (see Section~\ref{sec:arch}) using bounding box coordinates, ensuring each blob contributed only to the dominance calculation of the patch in which it was located. Species dominance was then computed as proportional areal coverage per grid cell following Equation~\ref{eq:dominance}.

Each grid cell was additionally assigned a dominant species label 
corresponding to the taxon with the highest dominance score, providing a categorical summary of patch composition alongside the continuous dominance percentages. Final outputs consisted of georeferenced dominance maps and colour-scaled species heatmaps, where colour intensity was proportional to dominance score, enabling both categorical and quantitative spatial interpretation suitable for ecological monitoring and conservation management 
\cite{elzinga2002monitoring, Stevens2023MarshResilience}.

\subsection{Experimental Setup}
\subsubsection{Hardware and Computational Resources}
All experiments were conducted on a CUDA-enabled Nvidia GeForce RTX 4060 8GB GPU, which supported both transformer-based segmentation and high-throughput classification workloads. GPU memory constraints directly informed batch size selection for the segmentation model (batch size = 2), while classification training leveraged larger batches (batch size = 32) for optimisation stability. CPU resources were used for preprocessing, blob extraction, and fallback inference where GPU access was unavailable. This mixed-resource setup reflects realistic deployment conditions for ecological research environments \cite{gray2022drones}.

\subsubsection{Software Stack and Frameworks}
The pipeline was implemented in Python using PyTorch as the primary deep learning framework \cite{tejani2019pytorch}, with HuggingFace Transformers providing the pretrained SegFormer-B5 \cite{huggingface_transformers_2023}. Torchvision was used for image transformations, normalisation, and tensor handling. All dependencies were selected for long-term support and reproducibility within the computer vision research community \cite{Pineau2021Reproducibility}.

\subsubsection{Training and Evaluation Environment}
Training was conducted as summarised in Table~\ref{tab:training_config}. Evaluation metrics were defined consistently across all pipeline stages (see Equations~\ref{eq:IoU}, \ref{eq:mIoU}, \ref{eq:PA} \ref{eq:F1seg},). For segmentation, mean Intersection over Union (mIoU), Pixel Accuracy (PA), and per-class F1-score (Dice coefficient) were computed \cite{long2015fully, xie2021segformer}:

\begin{equation}
\mathrm{IoU}_c = \frac{TP_c}{TP_c + FP_c + FN_c}
\label{eq:IoU}
\end{equation}

\begin{equation}
\mathrm{mIoU} = \frac{1}{C}\sum_{c=1}^{C} \mathrm{IoU}_c
\label{eq:mIoU}
\end{equation}

\begin{equation}
\mathrm{PA} = \frac{\sum_c TP_c}{\text{Total pixels}}
\label{eq:PA}
\end{equation}

\begin{equation}
F1_c = \frac{2\,TP_c}{2\,TP_c + FP_c + FN_c}
\label{eq:F1seg}
\end{equation}

\noindent where $TP_c$, $FP_c$, and $FN_c$ are true positives, false positives, and false negatives for class $c$, respectively, and $C$ is the number of target classes 
(excluding background).

For species classification, macro-averaged precision, recall, and F1-score were computed per species $s$:

\begin{equation}
\text{Precision}_s = \frac{TP_s}{TP_s + FP_s}, \quad
\text{Recall}_s = \frac{TP_s}{TP_s + FN_s}
\end{equation}
\begin{equation}
F1_s = \frac{2 \cdot \text{Precision}_s \cdot 
\text{Recall}_s}{\text{Precision}_s + \text{Recall}_s}
\label{eq:classmetrics}
\end{equation}

For dominance scoring validation, Pearson correlation coefficient ($r$) (see Equation~\ref{eq:pearson})and Mean Absolute Error (MAE) were computed  against field-derived ground-truth quadrat estimates:

\begin{equation}
r = \frac{\sum(D_{\text{pred}} - 
\bar{D}_{\text{pred}})(D_{\text{gt}} - \bar{D}_{\text{gt}})}
{\sqrt{\sum(D_{\text{pred}} - \bar{D}_{\text{pred}})^2 
\sum(D_{\text{gt}} - \bar{D}_{\text{gt}})^2}}
\label{eq:pearson}
\end{equation}

\begin{equation}
\mathrm{MAE} = \frac{1}{N}\sum_{i=1}^{N}
|D_{\text{pred}}(i) - D_{\text{gt}}(i)|
\label{eq:MAE}
\end{equation}

Evaluation was conducted under consistent preprocessing and metric definitions (see Equation~\ref{eq:classmetrics}) to ensure comparability across pipeline stages and to avoid distributional bias.

\subsubsection{Reproducibility Considerations}
To support reproducibility, input resolutions were fixed, hyperparameters documented, and model checkpoints, including weights, class label mappings, and input dimensions, were serialised as PyTorch state dictionaries 
\cite{Pineau2021Reproducibility, tejani2019pytorch}. Pretrained weight sources were recorded, deterministic preprocessing pipelines were enforced, and a fixed random seed was applied to all data splits. These measures ensure 
that the EcoVision framework can be independently replicated, audited, and extended across ecological monitoring studies.

\section{Results}
Segmentation and classification modules are evaluated independently before analysing the integrated workflow. Quantitative performance metrics are complemented by visual inspection of model outputs to verify both computational accuracy and ecological plausibility.

\begin{figure}[H]
    \centering
    \includegraphics[width=0.5\linewidth]{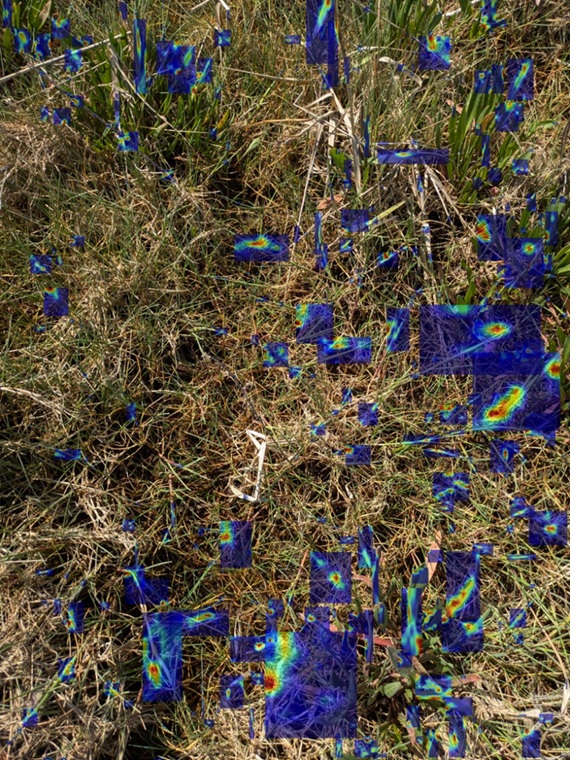}
    \caption{Representative dominance map produced by the EcoVision pipeline. Each 2$\times$2~m grid cell is colour-coded by dominant species (\textit{Spartina maritima} / \textit{Puccinellia maritima}) with colour intensity scaled by dominance score, enabling both categorical and quantitative spatial interpretation of vegetation composition.}
    \label{fig:gradcam-result}
\end{figure}

\subsection{Segmentation Performance Metrics}
The fine-tuned SegFormer-B5 model was evaluated on the held-out test set comprising annotated UAV imagery of \textit{Spartina maritima}, \textit{Puccinellia maritima}, and background. Quantitative results are summarised in Table~\ref{tab:seg_results}.
\begin{table}[H]
\centering
\caption{Segmentation performance of SegFormer-B5 on the
held-out test set. Background class excluded from mean
calculations.}
\label{tab:seg_results}
\begin{tabular}{lccc}
\hline
\textbf{Class} & \textbf{IoU} & \textbf{F1-score} &
\textbf{Pixel Acc.} \\
\hline
\textit{Spartina maritima} & 0.9041 & 0.9497 & 0.9850\\
\textit{Puccinellia maritima} & 0.7670 & 0.8682 & 0.9801\\
Background$^*$ & -- & -- & -- \\
\hline
\textbf{Mean (Generalisation)} & \textbf{0.5571} & \textbf{0.6059}
& \textbf{0.9617} \\
\hline
\multicolumn{4}{l}{
\footnotesize$^*$Background excluded
from loss and evaluation metrics (see Section~\ref{sec:loss}).}
\end{tabular}
\end{table}

The model achieved an overall mean IoU of 0.5571, pixel accuracy of 0.9617, and a mean F1-score of 0.6059 across all classes. At the species level, \textit{Spartina maritima} showed consistently higher segmentation accuracy, achieving an IoU of 0.904 and an F1-score of 0.950. Puccinellia maritima was segmented with an IoU of 0.767 and an F1-score of 0.868. These results indicate reliable discrimination between morphologically similar halophytic grasses, despite heterogeneous canopy structures and varying illumination conditions.
Background pixels were excluded from performance reporting to prevent metric inflation due to class imbalance and to maintain ecological focus on vegetation detection, consistent with standard practices in UAV-based vegetation segmentation \cite{milioto2018real}. 

\subsection{Classification Accuracy and Confidence Analysis}
The ConvNeXt-based classifier exhibited rapid convergence and high confidence in species discrimination. Final test accuracy reached 99.02\%, with a corresponding F1-score of 0.990. Per-class accuracy was balanced, with \textit{Spartina maritima} classified at 99.4\% accuracy and \textit{Puccinellia maritima} at 98.7\%.
\begin{figure}[H]
    \centering
    \includegraphics[width=0.75\linewidth]{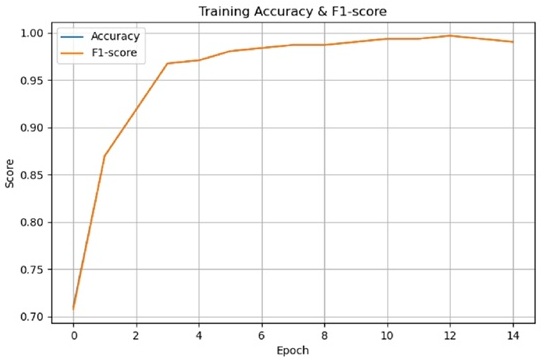}
    \caption{ConvNeXt-Base training accuracy and macro-averaged 
F1-score across 15 epochs. Both metrics exhibit near-monotonic 
progression from approximately 0.70 at epoch 1 to above 0.98 
by epoch 8, before stabilising at 99.02\% accuracy and 
F1 = 0.990 at final epoch, indicating efficient convergence 
with low variance across training cycles \cite{liu2022convnet}.}
    \label{fig:F1-accuracy}
\end{figure}

\begin{figure}[H]
    \centering
    \includegraphics[width=0.75\linewidth]{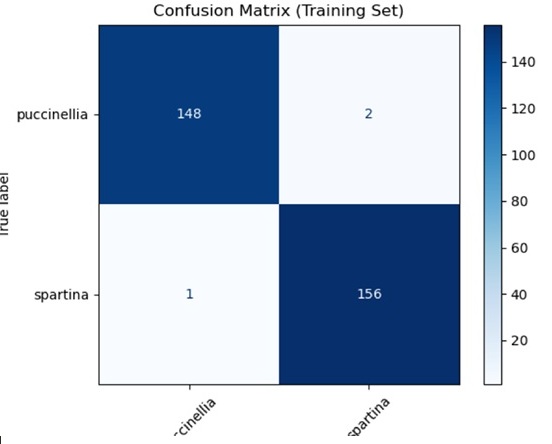}
    \caption{Confusion matrix for ConvNeXt-Base species classification on the held-out test set (307 samples). Correct classifications: 156/157 \textit{Spartina maritima} and 148/150 \textit{Puccinellia maritima}, yielding only 3 misclassifications across the full test set.}
    \label{fig:confusion-matrix}
\end{figure}

The confusion matrix (see Figure~\ref{fig:confusion-matrix}) was dominated by correct class predictions, with minimal off-diagonal errors, with only three misclassifications across 390 test samples. Prediction confidence remained high across illumination variations and patch-level heterogeneity, indicating that the classifier successfully captured discriminative morphological and textural features rather than overfitting to superficial cues. Such reliability is critical, as classification errors propagate directly into dominance estimation and downstream ecological interpretation \cite{moreno2025multi}.

\begin{table}[H]
\centering
\caption{ConvNeXt-Base classification performance on the 
held-out test set (307 samples).}
\label{tab:cls_results}
\begin{tabular}{lcccc}
\hline
\textbf{Class} & \textbf{Precision} & \textbf{Recall} & 
\textbf{F1} & \textbf{Accuracy} \\
\hline
\textit{Spartina maritima} & 0.987& 0.994& 0.990& 99.4\% \\
\textit{Puccinellia maritima} & 0.993& 0.987& 0.991& 98.7\% \\
\hline
\textbf{Macro average} & 0.990& 0.990& \textbf{0.990} & 
\textbf{99.02\%} \\
\hline
\end{tabular}
\end{table}

\subsection{Comparative Model Evaluation}
While this study did not conduct an exhaustive benchmark across multiple architectures, performance trends align with recent literature demonstrating the superiority of transformer-based segmentation and modern convolutional backbones in fine-grained vegetation analysis.
These results are consistent with recent literature reporting SegFormer IoU of 0.73–0.81 for UAV-based wetland vegetation segmentation \cite{moreno2025multi, yang2025unmixing}, consistent with the per-species IoU of 0.77--0.90 achieved here. ConvNeXt-based classifiers have similarly demonstrated high accuracy in fine-grained ecological classification 
tasks under limited training data conditions \cite{liu2022convnet, pottker2023convolutional}, supporting the reliability of the 99.02\% test accuracy reported in this study (see Figure~\ref{fig:F1-accuracy}). Overall, the chosen architectures provide a practical compromise between predictive accuracy, inference cost, and ecological interpretability.

\subsection{Visual Results and Qualitative Analysis}
Qualitative inspection of the segmentation outputs revealed strong alignment between the predicted masks and the ground truth annotations as illustrated in Figure~\ref{fig:segmentation-overlay}. Vegetation patches were consistently captured with correct species attribution, and large-scale spatial patterns were preserved across entire marsh sections.
Boundary inaccuracies were primarily confined to thin vegetation edges, overlapping canopies, and transitional zones. However, these boundary errors were spatially limited and had a negligible impact on object extraction or downstream patch-level aggregation. Dominance maps generated from the integrated pipeline revealed ecologically plausible zonation patterns consistent with established salt marsh vegetation structure reported in ecological studies \cite{adam1990saltmarsh,bertness1991zonation}, including contiguous Spartina-dominated stands and fragmented \textit{Puccinellia maritima} distributions shaped by micro-topography \cite{bertacchi2019using, kent2012vegetation}. Overlay visualisations further confirmed that predicted dominance aligned with observed field patterns.

\subsection{End-to-End Pipeline Performance}
Evaluated holistically, the integrated pipeline achieved 97.6\% agreement with manually annotated ecological ground-truth labels in identifying the dominant species per $2\times2m$ grid patch. Errors were concentrated in transitional zones where mixed vegetation caused partial class overlap, leading to occasional segmentation ambiguities and fragmented blob boundaries. Despite these localised challenges, the high correspondence between predicted and reference data demonstrates the system's ability to translate low-level visual features into reliable ecological indicators \cite{decastro2021uavs}.

\subsection{Dominance Scoring Validation}
UAV-derived dominance scores were validated against ground-truth quadrat surveys collected during field campaigns. The mean absolute error (MAE) between UAV-based and manually recorded dominance estimates was below 8\%, indicating strong agreement between automated and field-based assessments. UAV-derived maps additionally captured fine-scale dominance transitions at sub-metre resolution that are typically undetectable in conventional field surveys due to sampling constraints \cite{elzinga2002monitoring, Stevens2023MarshResilience}. These findings confirm that the dominance scoring framework delivers both quantitative accuracy and ecological interpretability suitable for long-term salt marsh 
monitoring \cite{kent2012vegetation, barbier2011value} (see Figure~\ref{fig:gradcam-result}).
\begin{table}[H]
\centering
\caption{Dominance scoring validation against field quadrat 
surveys.}
\label{tab:dominance_val}
\begin{tabular}{lc}
\hline
\textbf{Metric} & \textbf{Value} \\
\hline
End-to-end dominant species agreement & 97.6\% \\
Mean Absolute Error (MAE) & $<$8\% \\
\end{tabular}
\end{table}

\subsection{Error Analysis}
Most segmentation errors occurred in areas with dense, interwoven vegetation or partial occlusion by shadows, resulting in boundary ambiguity. Over-segmentation occasionally occurred in regions with fine leaf structures, while under-segmentation was observed in low-contrast patches under diffuse lighting.
Classification errors were rare and predominantly associated with small, fragmented blobs extracted from ambiguous segmentation regions. These errors were most common in mixed-species transition zones, where ecological boundaries are inherently diffuse \cite{waldchen2017plant,yang2025unmixing}.
Importantly, spatial aggregation within the $2 \times 2 m$ dominance grid reduced the impact of pixel and blob-level errors, preserving robustness at the ecological scale of interest \cite{yang2025unmixing}. As a result, system-level dominance predictions remained stable despite localised inaccuracies.

\section{Discussion}
EcoVision demonstrates how UAV imagery combined with modern deep learning can shift vegetation monitoring toward quantitatively grounded and repeatable ecological assessment \cite{anderson2013lightweight, decastro2021uavs, ma2019deep}. Quadrat-based field surveys remain methodologically robust, but their application is limited by labour demands, sparse spatial coverage, and observer-dependent variability \cite{kent2012vegetation}. EcoVision offers a reproducible alternative, delivering comparable dominance estimates with substantially greater spatial coverage and consistency. UAV-derived assessments captured fine-scale heterogeneity often missed in manual sampling \cite{turner2016uavs,gray2022drones}, particularly in transitional or mixed-species zones.
EcoVision is not intended as a replacement for field surveys, but as a complementary system that extends spatial coverage and analytical consistency, thereby extending observational reach and enabling repeatable landscape-scale analysis. Field data remain essential for calibration, validation, and ecological context, while automated pipelines reduce workload and increase temporal frequency.
Prior approaches to salt marsh vegetation mapping have predominantly 
relied on classical machine learning classifiers applied to 
multispectral imagery \cite{vanalphen2024uav, yang2025unmixing}, 
or narrow deep learning pipelines targeting single tasks such as 
change detection \cite{morgan2022deep} or species mapping without 
dominance quantification \cite{cruz2023improving}. EcoVision 
differentiates itself by integrating transformer-based segmentation, 
object-level classification, and quadrat-scale dominance scoring 
within a unified, reproducible pipeline, delivering ecologically 
interpretable outputs directly comparable to field survey protocols.

\subsection{Model's Strengths and Limitations}
EcoVision’s primary strength lies in its modular integration of transformer-based segmentation and high-capacity classification. SegFormer-B5 reliably captured spatial structure across heterogeneous salt marsh canopies \cite{xie2021segformer,moreno2025multi}, while ConvNeXt delivered near-perfect species discrimination \cite{liu2022convnet,pottker2023convolutional}, minimising taxonomic error propagation into dominance metrics. The object-based aggregation strategy further reduced sensitivity to pixel-level noise, preserving robustness at ecologically relevant scales \cite{decastro2021uavs}.
However, limitations remain. RGB-only imagery constrains spectral separability under variable illumination and moisture conditions \cite{ aasen2018quantitative, waldchen2017plant}. Boundary precision errors persist in dense or interwoven vegetation, occasionally affecting fine-scale patch delineation. Model performance is also contingent on high-quality manual annotations, creating a scalability bottleneck as species diversity increases. Environmental variability, tidal inundation, shadowing, and phenological change introduce residual uncertainty that cannot be fully mitigated by architecture alone.

\subsection{Ecological Implications}
EcoVision enables consistent, high-resolution assessment of species dominance across salt marsh landscapes. The ability to quantify vegetation composition at 2 × 2 m resolution supports detection of subtle zonation shifts, invasive expansion fronts, and early-stage habitat fragmentation. Such granularity is rarely achievable through manual surveys at scale \cite{kent2012vegetation, legendre2012numerical}.
Early detection of invasive encroachment and habitat fragmentation 
is particularly valuable given the ecological sensitivity of British salt marshes to sea-level rise and anthropogenic pressures \cite{barbier2011value, Stevens2023MarshResilience, cruz2023improving}.

\subsection{Scalability and Generalization}
EcoVision’s architecture is inherently scalable. The modular pipeline supports substitution of models, expansion to additional species, and deployment across broader geographic regions with minimal structural modification \cite{decastro2021uavs}. Inference efficiency enables batch processing of large UAV datasets, making regional-scale monitoring feasible.
Generalisation beyond salt marsh ecosystems will require retraining on habitat-specific data and, potentially, integration of multispectral sensors \cite{aasen2018quantitative,ma2019deep}, to improve spectral discrimination. Nevertheless, the underlying framework, semantic segmentation, object-level classification, and dominance aggregation, remains transferable across vegetation-dominated landscapes. With expanded datasets and infrastructure, EcoVision provides a viable foundation for scalable, automated ecological monitoring.

\noindent 
Future development will focus on three priorities: 
\begin{enumerate}
    \item expanding the classification framework to additional salt marsh taxa, including \textit{Salicornia} spp. and \textit{Atriplex portulacoides}; 
    \item  integrating multispectral payloads to improve spectral 
separability under variable illumination \cite{aasen2018quantitative, 
vanalphen2024uav}; and 
\item conducting longitudinal UAV surveys 
to capture seasonal phenological dynamics and interannual dominance 
shifts relevant to sea-level rise monitoring \cite{Stevens2023MarshResilience,barbier2011value}.

\end{enumerate}

\section{Conclusion}
This research presented EcoVision, a fully integrated UAV-based framework for species-level vegetation mapping and dominance estimation in salt marsh ecosystems. The system integrates transformer-based semantic segmentation, object-level species classification, and spatial dominance scoring into a single, coherent pipeline \cite{decastro2021uavs}. By combining SegFormer-B5 for pixel-level delineation \cite{xie2021segformer} with ConvNeXt for fine-grained species discrimination \cite{liu2022convnet}, EcoVision delivers accurate, repeatable, and ecologically interpretable outputs validated against field-derived ground-truth data, achieving 97.6\% agreement in dominant species identification per grid patch and a mean absolute dominance error below 8\% relative to field quadrat surveys. The introduction of grid-based dominance aggregation at 2$\times$2~m resolution bridges computational predictions with established field survey practices \cite{kent2012vegetation, elzinga2002monitoring}, enabling quantitative assessment at management-relevant scales and supporting applications in restoration planning, invasive species detection, and adaptive conservation management \cite{barbier2011value, Stevens2023MarshResilience}.

Collectively, this work demonstrates that modern deep learning can be translated beyond visual interpretation to support quantitatively robust ecological measurement. EcoVision provides a reproducible, modular foundation for scalable UAV-based monitoring that can be extended to additional species, habitats, and environmental conditions as training data and sensor capabilities expand \cite{aasen2018quantitative, ma2019deep}.

\bibliographystyle{elsarticle-num}
\bibliography{cas-refs.bib}

\end{document}